%% file: Main.tex
\documentclass{article}
\usepackage{microtype}
\usepackage{graphicx}
\usepackage{subfigure}
\usepackage{booktabs} 
\usepackage{subcaption}
\usepackage{caption}
\usepackage{hyperref}
\usepackage{amssymb}
\usepackage{amsmath}

\usepackage[accepted]{mlsys2025}

\begin{document}

\twocolumn[
\mlsystitle{Composition of Experts: A Modular Compound AI System Leveraging Large Language Models}




\begin{mlsysauthorlist}
\mlsysauthor{Swayambhoo Jain}{smb}
\mlsysauthor{Ravi Raju}{smb}
\mlsysauthor{Bo Li}{smb}
\mlsysauthor{Zoltan Csaki}{smb}
\mlsysauthor{Jonathan Li}{smb}
\mlsysauthor{Kaizhao Liang}{smb}
\mlsysauthor{Guoyao Feng}{smb}
\mlsysauthor{Urmish Thakkar}{smb}
\mlsysauthor{Anand Sampat}{smb}
\mlsysauthor{Raghu Prabhakar}{smb}
\mlsysauthor{Sumati Jairath}{smb}
\end{mlsysauthorlist}

\mlsysaffiliation{smb}{SambaNova Systems, Palo Alto, California, USA}

\mlsyscorrespondingauthor{Swayambhoo Jain}{swayambhoo.jain@gmail.com}
\mlsyscorrespondingauthor{Ravi Raju}{ravi.raju0594@gmail.com}

\vskip 0.3in

\begin{abstract}
Large Language Models (LLMs) have achieved remarkable advancements, but their monolithic nature presents challenges in terms of scalability, cost, and customization. This paper introduces the Composition of Experts (CoE), a modular compound AI system leveraging multiple expert LLMs. CoE leverages a router to dynamically select the most appropriate expert for a given input, enabling efficient utilization of resources and improved performance. We formulate the general problem of training a CoE and discuss inherent complexities associated with it. We propose a two-step routing approach to address these complexities that first uses a router to classify the input into distinct categories followed by a category-to-expert mapping to obtain desired experts. CoE offers a flexible and cost-effective solution to build compound AI systems. Our empirical evaluation demonstrates the effectiveness of CoE in achieving superior performance with reduced computational overhead.  Given that CoE comprises of many expert LLMs it has unique system requirements for cost-effective serving. We present an efficient implementation of CoE leveraging SambaNova SN40L \cite{prabhakar2024sambanova} RDUs unique three-tiered memory architecture. CoEs obtained using open weight LLMs \textit{Qwen/Qwen2-7B-Instruct, google/gemma-2-9b-it, google/gemma-2-27b-it, meta-llama/Llama-3.1-70B-Instruct and Qwen/Qwen2-72B-Instruct} achieve a score of $59.4$ with merely $31$ billion average active parameters on Arena-Hard \cite{Arena_Hard} and a score of $9.06$ with $54$ billion average active parameters on MT-Bench \cite{MT_Bench}.

\end{abstract}
]

\printAffiliationsAndNotice{}


\input{CoE_introduction}

\input{CoE_related_work}
\input{CoE_Notations}
\input{CoE_Formulation}

\input{CoE_Robust}
\input{CoE_Training}
\input{CoE_System_Considerations}
\input{CoE_Training_Data}
\input{CoE_Experiments_Updated}

\bibliography{CoE_references}
\bibliographystyle{mlsys2025}
\input{CoE_Appendix}
\end{document}

%% file: CoE_introduction.tex
\section{Introduction}

Large Language Models (LLMs) have significantly advanced the field of artificial intelligence. Modern LLMs benefit from the favorable scaling laws associated with transformer-based deep neural networks, which have shown no signs of performance saturation. This has led to the development of very large, trillion-parameter monolithic proprietary LLMs such as Gemini, GPT-4, and Claude. While these large monolithic models are justified by model scaling laws \cite{kaplan2020scaling}, they have drawbacks such as high serving costs, cumbersome maintenance and updates. Fine-tuning large language models presents significant barriers due to prohibitive computational costs that limit accessibility, and the inherent risk of task-specific specialization that can compromise model performance across broader domains due to the alignment tax \cite{ouyang2022training} and catastrophic forgetting \cite{luo2024empiricalstudycatastrophicforgetting}.

The recent release of open-weight LLMs like Llama3 \cite{touvron2023llama2}, Mistral \cite{jiang2023mistral}, and Gemma \cite{team2024gemma}, coupled with the active open-source LLM community, has democratized access to high quality pre-trained LLMs. This has resulted in a Cambrian explosion of diverse small-to-medium sized expert LLMs readily available on platforms like HuggingFace. These expert models address the long tail of needs arising from various domains, tasks, and languages that are poorly served by large monolithic LLMs. A narrow domain expert derived from a fine-tuned large monolithic LLM remains expensive to serve, making it economically impractical for catering to the long tail of niche requirements. Some of these expert models are known to outperform proprietary monolithic models on specific domains or tasks \cite{zhao2024lora, samba1}. In addition to this, natural variation in how popular general LLMs are trained they tend to have varying capabilities across different domains and languages. 

Motivated by the availability of many expert LLMs, in this paper we explore the problem of creating a compound AI system that combines several expert LLMs that caters to a practitioners application with a given parameter budget constraint. We propose a Composition of Experts (CoE), a compound AI system comprising a single router and multiple expert LLMs. The router's function is to dynamically select the most suitable expert for each incoming input prompt. If the router directs the given input to the correct expert, the CoE's overall performance on these tasks/domain will improve.

We begin by formulating the general problem of training a CoE, given training data, a set of experts, and a cumulative parameter budget. CoE  requires high-quality training data that allows to distinguish the capabilities of expert LLMs reliably. This is particularly challenging given that all LLM development follows benchmark driven approach. To address this challenge, motivated by recent work \cite{raju2024constructing} we follow a cost-effective semi-supervised approach to construct the training data for CoE. Another requirement to training such a CoE is the availability of labeled training data comprising of the best expert for each input. 

We demonstrate that labeling inputs with the best expert on a per-input basis results in a label distribution which has no discernible pattern that a router can exploit. To address this challenge, we propose a two-step routing approach. First, a category router maps the input to one of a finite set of categories. Second, a category-to-expert mapping assigns the input to a specific expert based on the category determined by the category router. We also provide a two-step training methodology to learn these mappings. While the category router training problem is a straightforward multi-class text classification task, the category-to-expert mapping is a more complex binary integer program. We reformulate this program to a tractable mixed-integer linear program (MILP), which can be solved using standard solvers available in standard scientific computing packages like SciPy \cite{Scipy}.

\begin{figure*}
    \centering
    \includegraphics[width=0.75\linewidth]{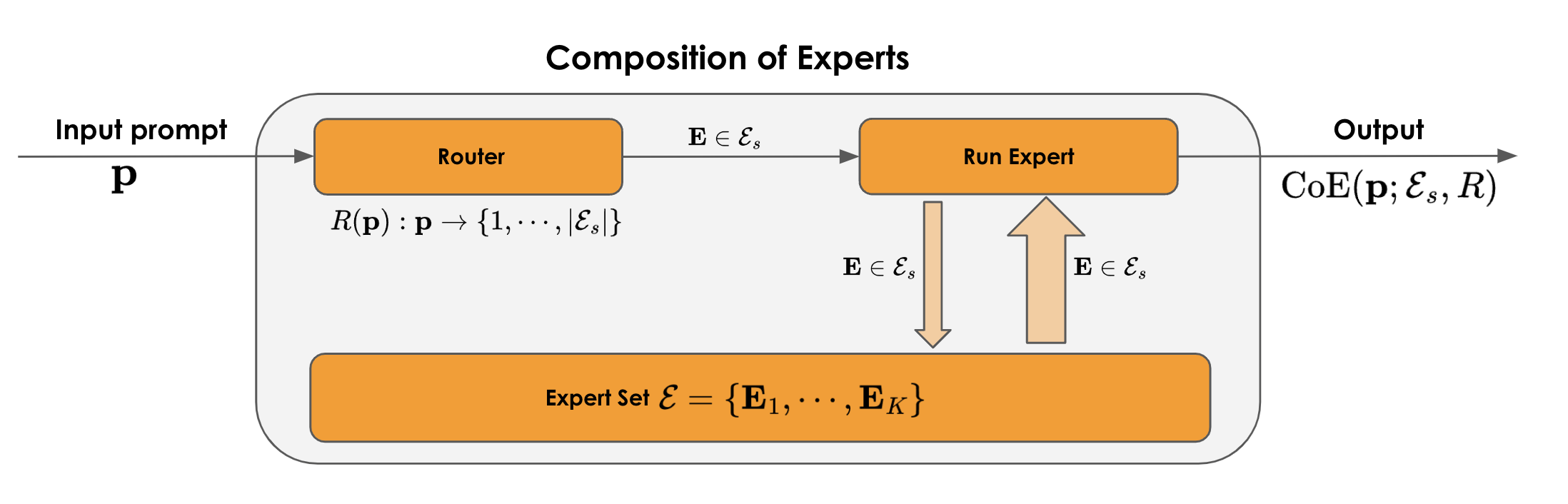}
    \caption{Abstract representation of Composition of Experts (CoE) for a given set of expert LLMs $\mathcal{E} = \{ \mathbf{E}_1, \cdots, \mathbf{E}_K\}$. For a given subset of experts $\mathcal{E}_s \subseteq \mathcal{E}$,  the $\textrm{CoE}(\mathbf{p};\mathcal{E}_s, R)$ routes the input prompt $\textbf{p}$ to one of the expert in $\mathcal{E}_s$ using the routing function $\textrm{R}(\mathbf{p})$ and produces the output by loading and the running that expert.} 
    \label{fig:CoE Basic System Diagram}
\end{figure*}

The input dependent dynamic expert selection in CoE has unique system requirements of large on-device memory to store the all the experts with sufficient enough bandwidth for quick model switching. We discuss limitations of naive implementation with mainstream GPUs and present an efficient implementation of CoE leveraging SambaNova SN40L \cite{prabhakar2024sambanova} RDU's unique three-tiered memory architecture.

The two step routing endows CoE with much needed modularity that has several advantages over traditional, monolithic LLMs. Unlike the time-consuming and resource-intensive processes of pre-training or fine-tuning large language models, CoE can easily incorporate new features, such as coding or multilingual capabilities, by simply integrating an open-source module with expertise in that area. This modular approach not only reduces the computational overhead but also gives designers greater control and interpretability. For example, CoE enables designers to use analysis tools like t-SNE \cite{van2008visualizing} to identify and address system failures, whether they stem from mis-classifications by the router or suboptimal performance by a specific module.

In summary, the main contributions of this work are given below:
\begin{enumerate}
    \item We propose the CoE approach to build compound AI systems using several expert LLMs  and a router that chooses the best model for the given input prompt. We present several technical challenges associated with the generic problem of training CoE under a given cumulative parameter budget.
    
    \item We propose a simple two step routing approach based on first classifying input prompt into finite categories via a category router and choosing the best model for each category using the category-to-expert mapping. This routing approach endows the CoE with the desirable property of modularity, which makes this approach scalable to adding and removing experts.  

    \item We also propose Robust-CoE that leverages uncertainty quantified routing to reliably adapt to the new prompt workloads by effectively the distribution mismatch in the router training data.
    
    \item A two step approach to first train category router followed by training the category-to-expert mapping. We provide a novel re-formulation to train the category-to-expert mapping to a feasible MILP problem which can be solved via off-the-shelf solvers. 
    
    \item CoE requires the category router to have high confidence in the routing decision. To ensure this, we introduce a data pipeline which uses a semi-supervised approach to label a large corpus of unlabeled data which can be used as the training data for the router.

     \item We present an efficient implementation of CoE leveraging SambaNova SN40L RDU's unique three-tiered memory architecture.

    \item We demonstrate efficacy of CoE using open weight LLMs \textit{Qwen/Qwen2-7B-Instruct, google/gemma-2-9b-it, google/gemma-2-27b-it, meta-llama/Llama-3.1-70B-Instruct and Qwen/Qwen2-72B-Instruct} and text-embedding model {intfloat/e5-mistral-7b-instruct} based router. We provide positive empirical evidence towards the proposed approach via extensive evaluations. On Arena-Hard, CoE achieves a score of $59.4$ with $31$ billion average active parameters and on MT-Bench CoE achieves a score of $9.06$ with $53.87$ billion average active parameters.
    
\end{enumerate}

%% file: CoE_related_work.tex
\section{Related Work}
The Mixture of Experts (MoE) architecture based LLMs are a popular choice \cite{lepikhin2020gshard, jiang2024mixtral}. These LLMs are built using of MoE-layer that use a fine-grained routing by employing a gating function in each transformer module. In contrast, the proposed CoE approach leverages a more coarse grained input-based routing allowing an MoE-based LLM to be subsumed as one of the expert in the CoE. 

Authors in \cite{jang2023exploring} show that splitting tasks into separate expert LLMs instead of a single multitask LM for zero-shot inference can have dramatic increases in terms of accuracy. Branch-Tree-merge(BTM)/cBTM \cite{li2022branchtrainmerge, gururangan2023scaling} proposes to train several experts on domain-specific data from a seed model and then merging the experts together during inference. Along a similar vein, \cite{sukhbaatar2024branchtrainmix} use the Brain-Tree-mix algorithm to independently train experts from a seed model but then integrate the feed-forward layers into an Mixture-of-Experts style ensemble and average the self-attention layers. They also introduce a routing module implemented as a feed-forward layer with various mechanisms such as top-2 selection and finetune the entire model. The merging is mainly needed to reduce the cost of inference and this typically results in reduction in the models accuracy. The CoE approach stands apart from these approaches as it focuses on composing pre-existing LLMs without necessitating the training of expert models.

Model Ensembles have been extensively investigated historically \cite{kumar2004minimum, he2016deep, ju2018relative}, with renewed interest in the context of LLMs. LLM-Blender introduces a post-hoc ensemble learning method, which ranks and fuses outputs from multiple large language models (LLMs), employing a system named PairRanker to categorize outputs and then amalgamating the highest-rated outputs through a sequence-to-sequence model, GenFuser \cite{jiang2023llm}. \cite{lu2023routing} proposed an efficient method to route queries to the most knowledgeable LLM using reward-guided routing and tag-based label enhancement. These works require running a given input through all the LLMs, making them impractically resource-intensive at inference time. CoE addresses this inference-cost challenge by routing inputs to only one expert LLM, providing the same computation cost as naively querying a monolithic LLM. Recent work \cite{shnitzer2023llm} suggested the generation of a routing function by evaluating candidate LLMs against benchmark data. Compared to this, the proposed CoE approach focuses on composing pre-existing LLMs without necessitating reliance on prior benchmark data. 

%% file: CoE_Notations.tex
\section{Notation}
$\mathbf{x} \in \mathbb{R}^n$ denotes a $n$-dimensional vector with real entries and $\mathbf{b} \in \{0,1\}^n$ represents $n$-dimensional binary vector whose entries are either $0$ or $1$. $\mathbf{1}_n = [1,\cdots, 1]^T$ denotes the vector with all entries as $1$.
 $\mathbf{X} \in \mathbb{R}^{m \times n}$ denotes $m \times n$ matrix with real entries and $\mathbf{B} \in \{0,1\}^{m \times n}$ denotes matrix with binary entries. $\mathbf{I}_n$ denotes $n \times n$ identity matrix. For a given set $\mathcal{E}$ the number of elements in the set are denoted by  $|\mathcal{E}|$.

%% file: CoE_Formulation.tex
\section{Composition of Experts}
\begin{figure*}
    \centering
    \includegraphics[scale=0.375]{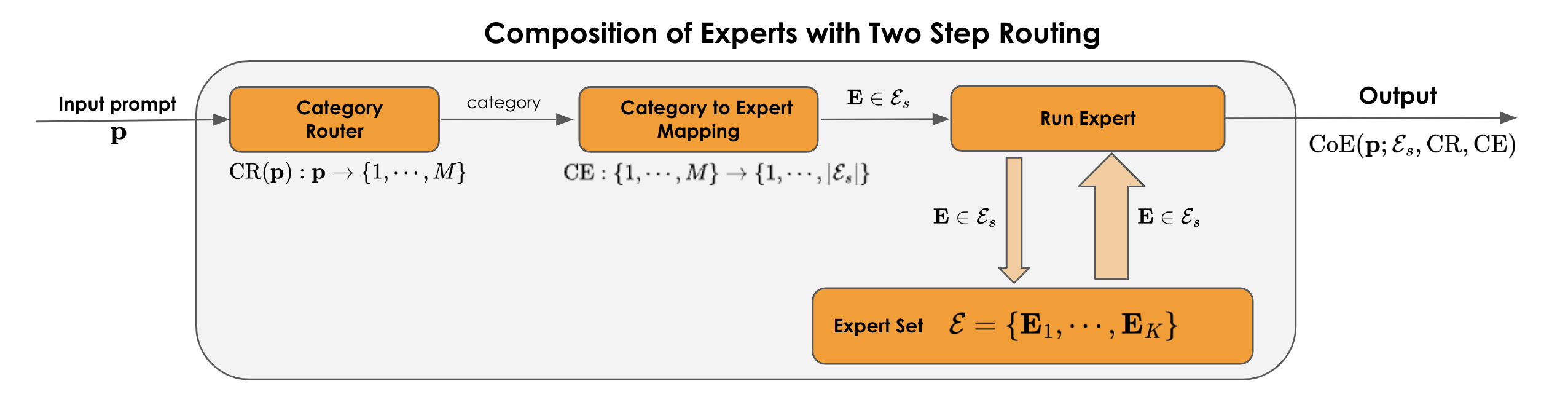}
    \caption{Abstract representation of Composition of Experts (CoE) with two step routing. For a given subset of experts $\mathcal{E}_s \subseteq \mathcal{E}$,  the $\textrm{CoE}(\mathbf{p};\mathcal{E}_s, \textrm{CE}, \textrm{CR})$ routes the input prompt $\textbf{p}$ by first mapping it to one of the $M$ categories using category-router $\textrm{CR}(\mathbf{p})$ followed by category-to-expert mapping $\textrm{CE}\left( \textrm{CR}(\mathbf{p})\right)$ to choose a designated expert for that category.}
    \label{fig:two_step_router}
\end{figure*}
The Composition of Experts (CoE) is constructed from a subset of experts $\mathcal{E}_s$ chosen from set of experts LLMs $\mathcal{E} = \{ \mathbf{E}_1, \cdots, \mathbf{E}_K\}$ such that the given input prompt $\textbf{p}$ is first passed through a router $R(\mathbf{p})$ which maps input to \emph{one} of the expert LLMs in the subset and the output is produced by that LLM. Formally, the CoE is defined as
\begin{align}
    \textrm{CoE}(\mathbf{p};\mathcal{E}_s, R) = \sum_{\mathbf{E}_j \in \mathcal{E}_s} \mathbf{1}_{\{R(\mathbf{p}) = j\}} \mathbf{E}_j(\mathbf{p}),
    \label{eqn:CoE_definition}
\end{align}
where $\mathcal{E}_s \subseteq \mathcal{E}$,  the routing function is defined as $R(\mathbf{p}): \mathbf{p} \rightarrow  \{1, \cdots, |\mathcal{E}_s|\}$,  and $\mathbf{1}_{\{R(\mathbf{p}) = j\}}$ and is the canonical indicator function which evaluates to $1$ if $R(\mathbf{p}) = j $ and $0$ otherwise. The high level abstract representation of CoE is shown in Figure \ref{fig:CoE Basic System Diagram}. The cumulative number of parameter in $\mathcal{E}_s$ determines the inference cost for CoE. Furthurmore, it is desirable to construct CoEs under a given parameter budget $B$. Let $\mathcal{D} = \left\{(\mathbf{p}_1,\mathbf{c}_1) , \cdots, (\mathbf{p}_N, \mathbf{c}_N)\right\}$ be the training data comprising of $N$ prompt-completion pairs. The generic problem of training a CoE with training data $\mathcal{D}$, expert set $\mathcal{E}$ and parameter budget $B$ can be formally stated as follows:  
\begin{equation}
    \begin{aligned}
     \min_{ \mathcal{E}_s \subseteq \mathcal{E} } &  \sum_{ (\mathbf{p}_i,\mathbf{c}_i) \in \mathcal{D}} \mathcal{L} \left( \textrm{CoE}(\mathbf{p};\mathcal{E}_s, R), \mathbf{c}_i   \right) \ \\ 
     \textrm{ s.t. } & \sum_{\mathbf{E}_i \in  \mathcal{E}_s } |\mathbf{E}_i| \le B, 
        \label{eqn:base_problem}
    \end{aligned}
\end{equation}
where $|\mathbf{E}_i|$ denote the total parameters in LLM $\mathbf{E}_i$ and $\mathcal{L}(\cdot)$ is the desired loss function capturing quality of completions produced by CoE. This formulation is quite generic as the loss function could be as simple as likelihood of generating $\mathbf{c}_i$ corresponding to $\mathbf{p}_i$ or it could be LLM as a judge score given by a monolithic large LLM. 
\begin{figure}
    \centering
    \includegraphics[width=1\linewidth]{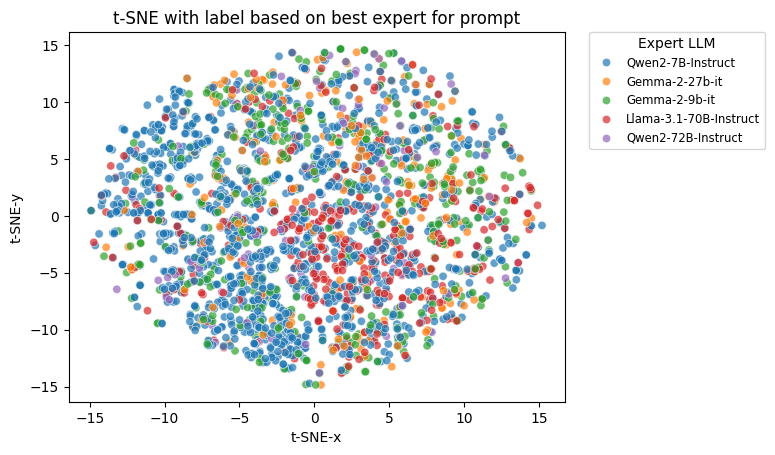}
    \caption{2D t-SNE plot for prompt-embeddings obtained from text-embedding model \textit{intfloat/e5-mistral-7b-instruct} for prompts in the CoE training data. Labels based on best expert LLMs chosen from the expert set comprising of \textit{Qwen/Qwen2-7B-Instruct, google/gemma-2-9b-it, google/gemma-2-27b-it, meta-llama/Llama-3.1-70B-Instruct and Qwen/Qwen2-72B-Instruct}. Best expert is obtained by using LLM-as-a-judge with details provided in Section \ref{sec:experimental_setup}. }
    \label{fig:2d_tsne_plot_router_label}
\end{figure}

Given the non-differentiable nature of the indicator function $\mathbf{1}_{\{R(\mathbf{p}) = j\}}$, training CoE via solving the general problem in \eqref{eqn:base_problem} is non-tractable. To this end, an intuitive approach would be to label the each prompt with the best expert in expert subset $\mathcal{E}_s$ satisfying the given parameter budget $B$ generate completions for each expert and obtain routing labels based on expert that incurs minimum loss w.r.t to desired completion. Equipped with labeled routing data, one can train a multi-class text classifier that maps each input to best expert.  Even if we ignore the combinatorial nature of parameter budget constraint this approach has a fundamental issue related to the lack of pattern for the router to learn. Similar prompts have completely different experts and labeling based on best expert is too jittery.  In Figure \ref{fig:2d_tsne_plot_router_label}, we show the 2D t-SNE plot for the prompts in the training data where the colors indicate the best expert model for each prompt. Best expert is obtained by using LLM-as-a-judge with details provided in Section \ref{sec:experimental_setup}. We can clearly see that there is no discernible pattern and a router trained with this data would be highly inaccurate. We experimented with a variety of distance metrics between the desired completions and expert completion to get the best expert. The labeling distribution was consistently jittery. A possible explanation for this can be the overlap in the training data of the experts.

\subsection{CoE with Two Step Routing}
\begin{figure}
     \centering
     \includegraphics[width=0.90\linewidth]{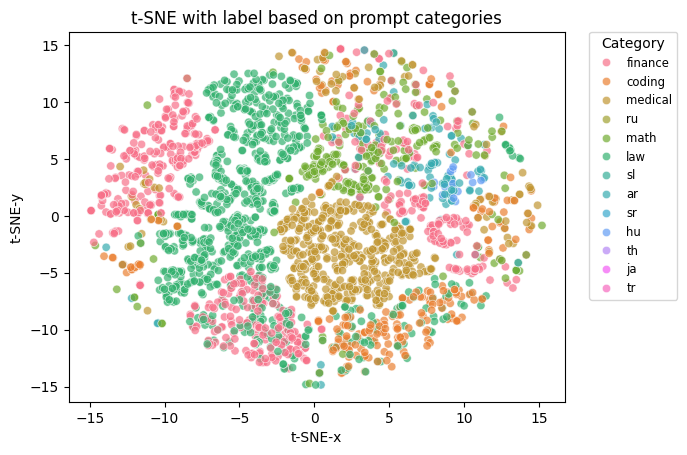}
     \caption{2D t-SNE plot for prompt-embeddings obtained from text-embedding model \textit{intfloat/e5-mistral-7b-instruct} for prompts in the CoE training data.  Labels are based on various categories comprising of variety of domains and languages.
    }
    \label{fig:2d_tsne_plot_category}
\end{figure}

While directly labeling each prompt for best expert is difficult to train a accurate router. We observe in Figure \ref{fig:2d_tsne_plot_category} that when prompts are labeled based on natural categories such as medical, coding, etc. we see clear clusters. Motivated by this natural clustering of prompts in distinct categories, a reasonable approach is to assign best expert for each category. Since there are many prompts in each category, it doesn't suffer from labeling jitteriness as we encountered while assigning best expert on per-prompt basis.

Based on this, we impose a simplifying constraint on the router such that the input prompt $\mathbf{p}$ is first mapped to one of $M$ categories and subsequently followed by routing it to one of the expert for that category using with the category-to-expert mapping. More concretely, the router is a composite mapping as $R(\mathbf{p}) = \textrm{CE}( \textrm{CR}(\mathbf{p}))$ where $\textrm{CR}: \mathbf{p} \longrightarrow \{1,\cdots,M\}$ is the category-router that maps input prompt $\mathbf{p}$ to one of the $M$ categories and $\textrm{CE}:\{1,\cdots,M\} \longrightarrow \{1,\cdots,|\mathcal{E}_s|\} $ is the category-to-expert mapping. Formally, the CoE under this two-step router can be mathematically stated as
\begin{align}
    \textrm{CoE}(\mathbf{p};\mathcal{E}_s, \textrm{CE}, \textrm{CR}) & = \sum_{\mathbf{E}_j \in \mathcal{E}_s} \mathbf{1}_{\{\textrm{CE}( \textrm{CR}(\mathbf{p}) ) = j\}} \mathbf{E}_j(\mathbf{p})
        \label{eqn:CoE_two_step_definition}
\end{align} 
The high level system diagram for CoE with two step routing approach is shown in Figure \ref{fig:two_step_router}. Under this setup, the category router $\textrm{CR}$ can be a simple multi-class text classifier trained to identify categories and category-to-expert mapping can be obtained by solving resulting optimizing problem for given  $\textrm{CR}$. This two step approach provides a nice modularity to routing wherein expert set and categories can be updated in a modular fashion. More details are provided in Section \ref{sec:training_algo}. 


%% file: CoE_Robust.tex
\subsection{Robust-CoE with Uncertainty Quantified Routing}
For extending CoE generalization beyond the labeled category routing training data we propose Robust-CoE with uncertainty quantified routing.  
The uncertainty quantification has been exhaustively studied in the active learning literature \cite{settles2009active, rakesh2021efficacy} and adversarial robustness in deep learning \cite{goodfellow2018making, sheikholeslami2020minimum, sheikholeslami2019efficient}. A variety of metrics such as such as entropy, variance ratio etc. on the probabilities the category-router assigns to various categories for a given text input can be used as a measure of uncertainty.
If the uncertainty is higher than a given threshold then we assign the prompt to 'general' category. The two step routing approach can be easily adopted for this wherein the category router outputs $M+1$ categories containing 'general' as one of the category. 

%% file: CoE_Training.tex
\section{Training Algorithm for CoE}\label{sec:training_algo}
For training the two-step routing CoE, we propose to first train the category-router, $\textrm{CR}$ followed by training the category-to-expert mapping $\textrm{CE}$.
\subsection{Step 1: Train the category router $\textrm{CR}$}
Training the category router requires labeled training data $\mathcal{D}^{CR} = \{(\mathbf{p}_1, r_1), \cdots, (\mathbf{p}_N, r_N)\}$  where $r_i$ is the category label for prompt $\mathbf{p}_i$. Equipped with $\mathcal{D}^{CR}$, we can train a multi-class text classifier to obtain $\hat{\textrm{CR}}$. A strong pre-trained text-embedding model and a light-weight multi-class classifier built on top of these text-embedding model is an effective choice for the category router architecture. 

\subsection{Step 2: Train the category-to-expert mapping CE}
We substitute the category router $\hat{\textrm{CR}}$ from Step $1$ in the generic problem in \eqref{eqn:CoE_two_step_definition} to obtain the following optimization problem for learning category-to-expert mapping $\textrm{CE}$ 
\begin{equation}
    \begin{aligned}
     \min_{ \mathcal{E}_s \subseteq \mathcal{E} }  & \sum_{ (\mathbf{p}_i,\mathbf{c}_i) \in \mathcal{D}} \mathcal{L} \left( \textrm{CoE}(\mathbf{p};\mathcal{E}_s, \textrm{CE}, \hat{\textrm{CR}}), \mathbf{c}_i   \right) \\
     \textrm{ s.t. } &  \sum_{E_i \in  \mathcal{E}_s } |\mathbf{E}_i| \le B 
        \label{eqn:CoE_problem}
    \end{aligned}
\end{equation}
The above problem can be converted to a well studied mixed integer linear program using the following facts:

\begin{itemize}
    \item The category router $\hat{\textrm{CR}}$ partitions $\mathcal{D}$ into $M$ parts $\mathcal{D}_1,\cdots, \mathcal{D}_M$, where $\mathcal{D}_m$ is the set of prompts in training data that are mapped to category $m$. 
    \item Let $l_{ij}$ denote the cost of mapping prompts in category $i$ to expert $j$ and in terms of $\mathcal{D}_m$ this cost is denoted by 
    \begin{align*}
          l_{ij} = \sum_{ (\mathbf{p}_k, \mathbf{c}_k) \in \mathcal{D}_i} \mathcal{L} \left(\mathbf{E}_j(\mathbf{p}_k), \mathbf{c}_k   \right) 
    \end{align*}
    \item $\textrm{CE}$ is a mapping from $\{1, \cdots, M\}$ to $\{1, \cdots, K\}$ it can  be represented as binary matrix $\mathbf{C} \in \{0,1\}^{M \times K}$. Since each category is served by only one expert model, the rows of $\mathbf{C}$ are constrained to have only $1$ non-zero entry in each row, i.e., it must satisfy $\mathbf{C}\mathbf{1}_K = \mathbf{1}_M$. 

    \item In terms of $\mathbf{C}$, the set of experts $\mathcal{E}_s$ can be represented as the set of non-zero indices of the vector  $\mathbf{C}^T\mathbf{1}_M$. Let $\textrm{supp}(\mathbf{C}^T\mathbf{1}_M)$ be the binary vector that is value $1$ only at indices where $\mathbf{C}^T\mathbf{1}_M$ is non-zero.

    \item Let's also define the expert size vector $\mathbf{s}_{\mathcal{E}}  = [|\mathbf{E}_1|, \cdots,|\mathbf{E}_K|]^T$. Then the total number of parameters in the CoE for a given $\mathbf{C}$ can be written as, $\mathbf{s}_{\mathcal{E}}^T \textrm{supp}(\mathbf{C}^T\mathbf{1}_M)$.
    


\end{itemize}

\begin{figure*}
    \centering
    \includegraphics[width=0.60\linewidth]{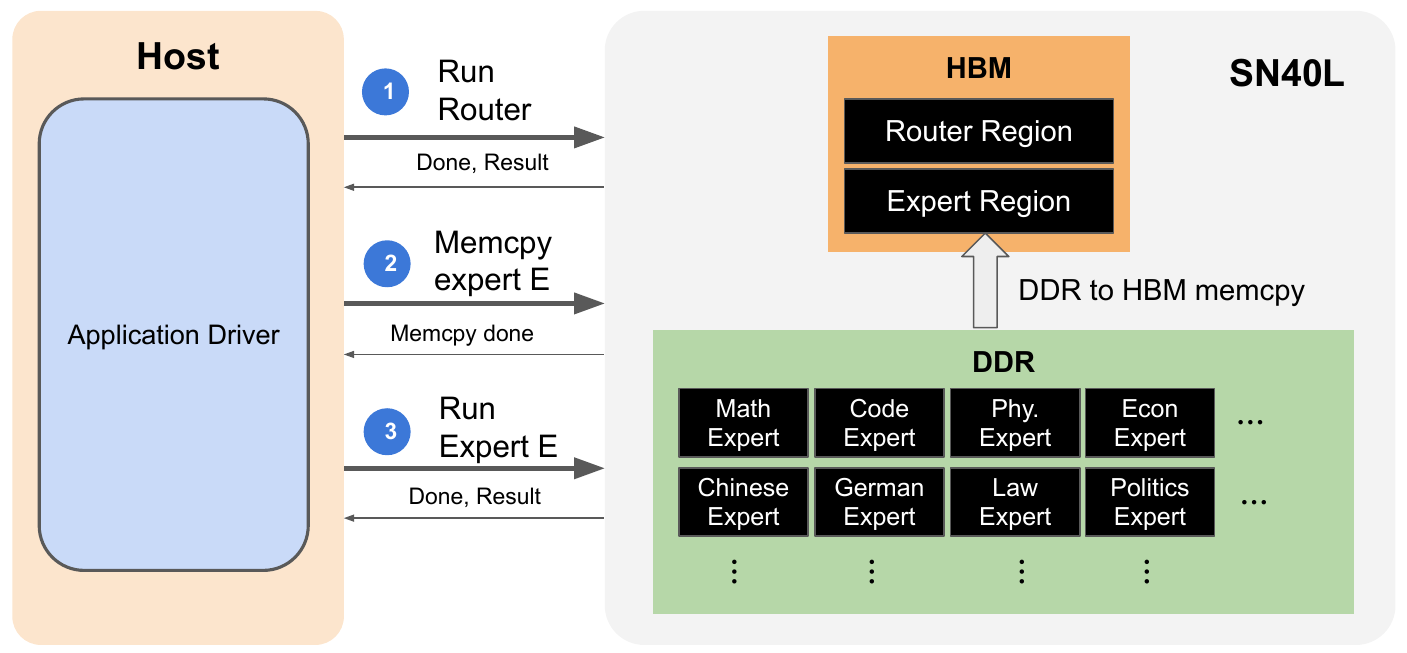}
    \caption{A simplified sequence of operations for CoE serving via SN40L. Router weights are in HBM. Expert weights are in DDR, with a region pre-allocated in HBM for the ``current" expert(s).}
    \label{fig:SN40L_Serving}
\end{figure*}

Equipped with above facts, the problem \eqref{eqn:CoE_problem} for a given category router $\hat{\textrm{CR}}$ can be re-written in the following equivalent form:
\begin{equation}
\begin{aligned}
 \min_{\mathbf{C}} & \ \sum_{i,j} l_{ij} c_{ij}\\
\textrm{s.t.}  & \   \mathbf{s}_{\mathcal{E}}^T \textrm{supp}(\mathbf{C}^T\mathbf{1}_M) \le B,\\
&  \mathbf{C} \in \{0,1\}^{M \times K}, \mathbf{C}\mathbf{1}_K = \mathbf{1}_M,
\end{aligned}
\end{equation}
where the constraint $\mathbf{s}_{\mathcal{E}}^T \textrm{supp}(\mathbf{C}^T\mathbf{1}_M) \le B$ ensures the parameter budget is satisfied.

Next, we show that the term $\textrm{supp}(\mathbf{C}^T\mathbf{1}_M)$ can be replaced by an equivalent binary optimization variable $\mathbf{y} \in \{0,1\}^K$ with the constraints $\mathbf{C}^T \mathbf{1}_M \le M \mathbf{y}$ and $\mathbf{y} \le \mathbf{C}^T \mathbf{1}_M$. The argument goes as follows: If any element in the $i^{th}$ column of $\mathbf{C}$ is nonzero then the $i^{th}$ entry of $\mathbf{y}$ is always $1$ because of $ \mathbf{C}^T \mathbf{1}_M \le M \mathbf{y}$ and the condition  $\mathbf{y} \le \mathbf{C}^T \mathbf{1}_M$ is automatically satisfied. When all the entries of the $i^{th}$ column of $\mathbf{C}$ are zero then the constraint $\mathbf{y} \le  \mathbf{C}^T\mathbf{1}_M $ will force $\mathbf{y}$ to be zero and $ \mathbf{C}^T \mathbf{1}_M \le M \mathbf{y}$ is automatically satisfied. Simply stated the binary variable $\mathbf{y}$ along with the constraint $\mathbf{y} \le \mathbf{C}^T \mathbf{1}_M \le M \mathbf{y}$ encodes the indices of experts chosen in CoE via its non-zero entries and the corresponding term $\mathbf{s}_{\mathcal{E}}^T \mathbf{y}$ is equal to the total number of parameters in the corresponding CoE. Equipped with this for solving category-to-expert mapping we need to solve the following mixed integer linear program
\begin{equation}
    \begin{aligned}
 \min_{\mathbf{C}, \mathbf{y}} & \ \sum_{i,j} l_{ij} c_{ij}\\
\textrm{s.t.}  & \  \mathbf{C} \in \{0,1\}^{M \times K}, \mathbf{y} \in \{0,1\}^K,\\
&  \mathbf{s}_{\mathcal{E}}^T \mathbf{y}  \le B, \mathbf{y} \le \mathbf{C}^T \mathbf{1}_M \le M \mathbf{y}, \\
&\mathbf{C} \mathbf{1}_K = \mathbf{1}_M. 
\label{eqn:MILP_Complicated}
\end{aligned}
\end{equation}
This problem can be directly fed into any off-the-shelf freely available standard MILP solvers. This requires vectorization of the matrix variables by stacking columns on top of each other by using the mathematical identity $\textrm{Vec}(\mathbf{AXB}) = \left(\mathbf{B}^T \otimes \mathbf{A} \right)\textrm{Vec}(\mathbf{X})$ for vectorization. With this \eqref{eqn:MILP_Complicated} in standard MILP format is given by
\begin{equation}
    \begin{aligned}
 \min_{\mathbf{c}, \mathbf{y}} & \ \mathbf{c}^T\mathbf{l}\\
\textrm{s.t.}  & \ \mathbf{c} \in \{0,1\}^{MK},  \mathbf{y} \in \{0,1\}^K, \\
& \begin{bmatrix}
   \mathbf{0}_{MK}^T & \mathbf{s}_{\mathcal{E}}^T \\
   \mathbf{I}_K \otimes \mathbf{1}_M^T & -M\mathbf{I}_K \\
   -\mathbf{I}_K \otimes \mathbf{1}_M^T & \mathbf{I}_K  \\
\end{bmatrix} \begin{bmatrix}
     \mathbf{c} \\
     \mathbf{y}
\end{bmatrix} \le  \begin{bmatrix}  B \\ \mathbf{0}_K \\ \mathbf{0}_K  \end{bmatrix}, \\
& \begin{bmatrix} \mathbf{1}^T_K \otimes \mathbf{I}_M &\mathbf{0}_{M \times K} \end{bmatrix} \begin{bmatrix}
     \mathbf{c} \\
     \mathbf{y}
\end{bmatrix} = \mathbf{1}_M. \label{eq: MILP}
\end{aligned}
\end{equation}

The above two step approach on a given datasets $\mathcal{D}, \mathcal{D}^{CR}$ enables a modular learning process thereby making adding/removal of categories and experts to existing CoE a tractable process. Such modularity also allows for reduced inference cost by choosing smaller experts, interpretability, etc.  The two step training algorithm can be easily modified to train Robust-CoE by simply computing win-rates of experts on the set of prompts that are assigned  to `general' category and obtaining the category-to-expert by solving resulting optimization problem in \eqref{eq: MILP}.

%% file: CoE_System_Considerations.tex
\section{System considerations for CoE} 
CoE provides the capability to improve the model incrementally by adding more experts in a modular fashion. While the total number of parameters in CoE can increase, the number of parameters required to serve any given input query stays bounded. Specifically, each input query would require the parameters for the router and one of the experts. Theoretically, this property enables sustaining the same response throughput even as the number of experts increase.

\begin{figure*}
    \centering
    \includegraphics[scale=0.225]{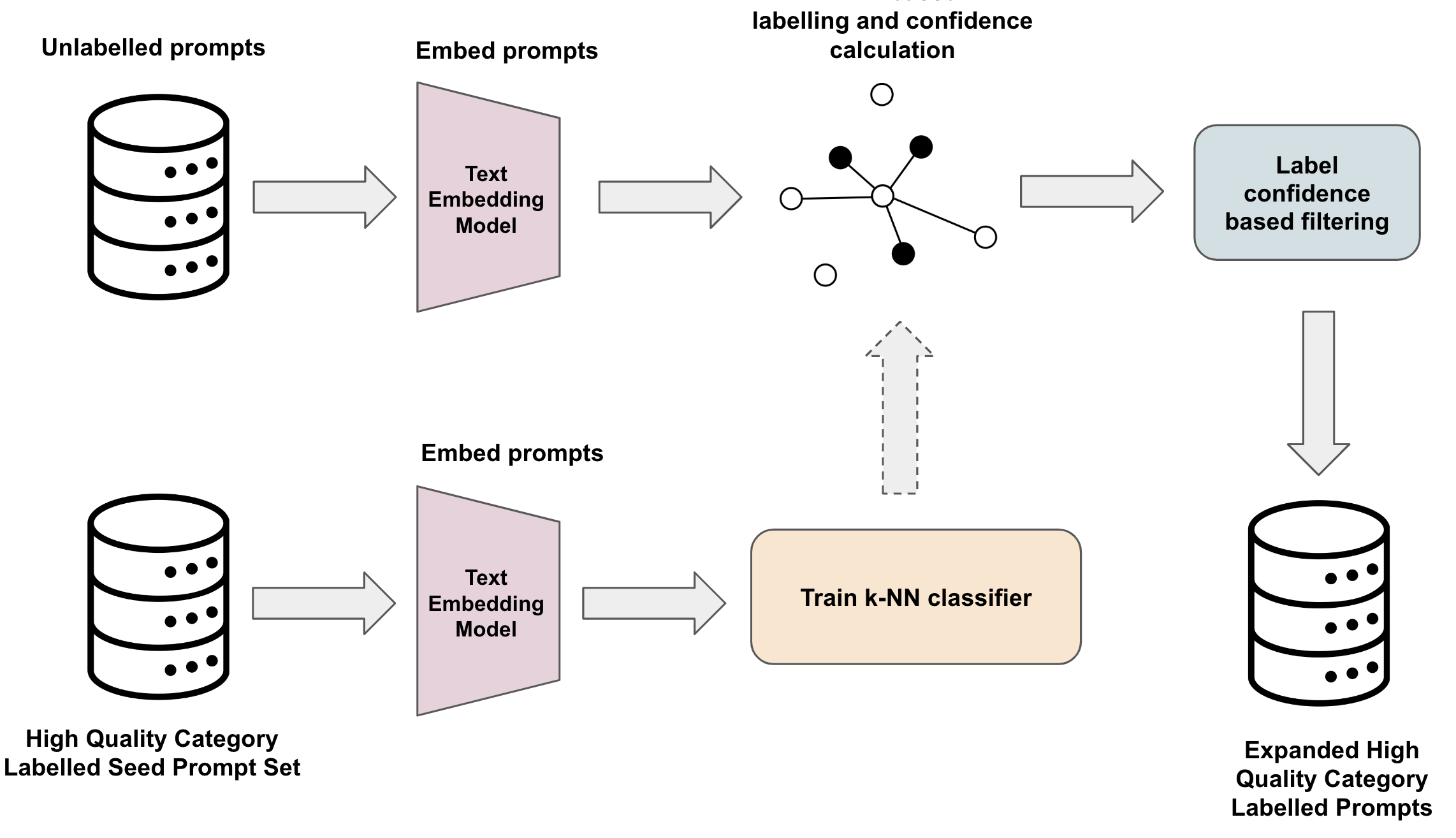}
    \caption{Semi-supervised pipeline to curate high quality prompts for CoE training data for various categories. Starting with high-quality labeled small seed dataset we leverage text-embeddings to train a $k$-NN classifier. The $k$-NN classifier is used to further expand the training by labeling new prompt with categories and filtering based on labelling confidence.}
    \label{fig:training_data_pipeline}
\end{figure*}

However, a naive implementation of CoE quickly runs into memory capacity limits on GPUs. Modern H100 GPUs are equipped with about 80 GB of High-Bandwidth Memory (HBM) per GPU socket~\cite{dgx-h100}. Model parameters are typically loaded once into HBM during initialization. Incoming queries are then executed on the GPU with the model parameters and KV cache served from HBM memory. We can see that the HBM capacity utilized depends on the number of parameters and KV cache size, where the latter scales linearly with both the batch size and sequence length. Consequently, large CoE configurations can run out of HBM capacity. Capacity issues can be mitigated in two ways: (a) Use more GPUs to obtain higher net HBM capacity, or (b) Use the host CPU's DRAM as an overflow space. Using more GPUs is expensive. For instance, a CoE with 150 8B experts can require up to 19 DGX H100 systems~\cite{prabhakar2024sambanova, sn40l-hotchips}. Furthermore, as the traffic distribution to each expert can fluctuate over time, using multiple GPUs in an ``expert parallel" manner can create bottlenecks with expert hotspots, where several requests route to only a subset of GPUs containing the popular expert. On the other hand, using the CPU's DRAM as overflow creates a bottleneck when models need to be copied to GPU HBM. For instance, a DGX H100 provides a CPU-to-GPU bandwidth of 64 GB/s~\cite{dgx-h100}. At this bandwidth, copying the weights of a single 8B expert would optimistically take 233 milliseconds. Considering that models like Llama3.1 8B run at over 300 tokens/s on H100~\cite{aa}, or just 3 milliseconds per token, such high model copy costs lowers the overall performance of the system.

To mitigate the issues above, we first observe that CoE exposes data locality at the expert level that is best exploited with a second memory tier. The sequence of operation are shown in Figure \ref{fig:SN40L_Serving}. Storing all parameters and KV caches requires a memory with a large capacity, but also a reasonably large bandwidth to speed up model copy costs. Running inference on an expert requires small but high bandwidth memory to exploit the reuse in parameters when multiple tokens are generated in an autoregressive manner. New AI accelerator systems like the SambaNova SN40L~\cite{prabhakar2024sambanova, sn40l-hotchips} and NVIDIA Grace Hopper~\cite{gh200} are architected with a tiered memory system with terabytes of high-capacity DDR memory as well as gigabytes of high-bandwidth HBM memory. The existence of both DDR and HBM enables cost-efficient CoE inference by reducing the machine footprint to host model parameters, and also by minimizing model switching overheads. SambaNova SN40L, for instance, has 12 TB of DDR, providing storage for about 5 trillion parameters~\cite{prabhakar2024sambanova}. With an aggregate bandwidth of 800 GB/s, copying a Llama 3.1 8B from DDR to HBM would be completed in 19 milliseconds. The SN40L has higher aggregate DDR capacity, and has demonstrated 3$\times$ higher token generation throughput over GPUs. Consequently, we adopt and evaluate CoE on the SambaNova SN40L in this paper.


%% file: CoE_Training_Data.tex
\section{Semi-Supervised Training Data Pipeline For CoE}
In order to fully realize the potential of CoE, high quality prompts in the training data are needed so that the router can distinguish between experts LLMs on variety of domains and languages.  Given the competitive landscape of LLMs and most of the LLMs are optimizing performance on the same benchmarks. This makes distinguishing LLM capabilities based on existing benchmarks a challenging problem because all the models perform similarly. Getting high quality prompts that allow us to distinguish between different LLM is cumbersome and costly. 

Recent work \cite{raju2024constructing} proposed a semi-supervised approach to build benchmarks with high degree of separability in the LLM performance in a cost effective manner. We leverage similar semi-supervised approach and use a prompt curation pipeline shown in Figure \ref{fig:training_data_pipeline}. Starting with a seed set comprising high quality category labeled prompts we first train a $k-$NN based multi-class text classifier that uses text-embedding from a pre-trained text embedding model. 

Equipped with this $k$-NN classifier we label a much larger unlabeled prompt. For quality control we also measure the classifier confidence on the $k$-NN labels and filter out low confidence prompts. The threshold used for confidence based filtering can be adjusted to strike a balance between quality and quanity of prompts obtained after filtering. 

%% file: CoE_Experiments_Updated.tex
\section{Experimental Setup}\label{sec:experimental_setup}
We evaluate the two-step routing based CoEs for the expert set $\mathcal{E}$  comprising of the following well known open weight LLMs: 
\textit{Qwen/Qwen2-7B-Instruct, google/gemma-2-9b-it, google/gemma-2-27b-it, meta-llama/Llama-3.1-70B-Instruct and Qwen/Qwen2-72B-Instruct}. We consider the following domain categories: Medical, Finance, Coding/Computer Programming, Mathematics, and Law; and the following languages categories: Arabic, Serbian, Slovenian, Hungarian, Russian, Turkish, Japanese and Thai.

\begin{figure}        
\centering
    \includegraphics[width=0.65\linewidth]{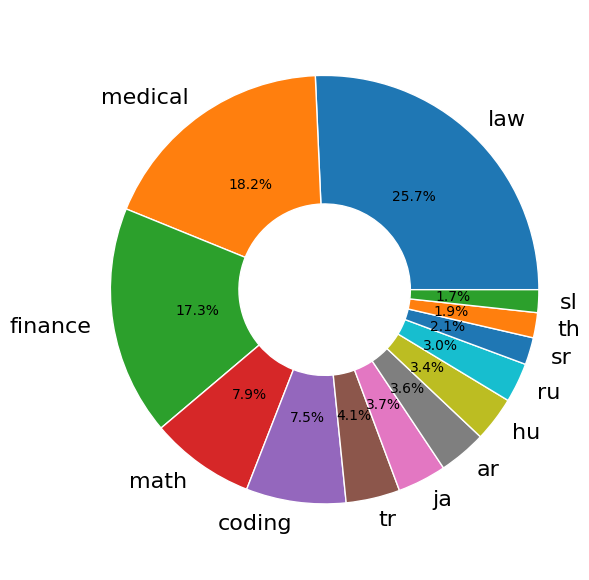}
    \caption{Prompt category distribution in the router training data.}
    \label{fig:router_cat_distribution}
\end{figure}

\subsection{CoE Training data}
 We source high quality seed prompts for these domain categories with domain specific open-source datasets available on HuggingFace. We restrict only to the training splits to avoid contamination with benchmarks. The dataset used for each domain is listed in Table \ref{tab:domain_dataset_sources} in the Appendix. We construct multilingual prompts in various formats such as native, translation, and cross-lingual for all the language categories. The exact methodology and source datasets are outlined in Section \ref{sec:multilingual_category} in the Appendix. 

Equipped with these high quality category labeled seed prompts extend prompt dataset by following the semi-supervised pipeline in Figure \ref{fig:training_data_pipeline} on general chat prompts from internal user trials and training split \cite{lmsys-chatbot-arena}. We use \textit{intfloat/e5-mistral-7b-instruct} as the text embedding model given its strong performance on Massive Multitask Embedding Benchmark \cite{muennighoff2023mtebmassivetextembedding} and train $k$-NN classifier on seed prompt set. We measure confidence in terms of entropy obtained by probabilities that the classifier assigns to various categories and we filter out prompts with entropy greater than $0.5$. The entropy threshold was chosen so that quality prompts with quantity were obtained. In this manner, curated about $58000$ high quality category labeled prompts.  
The Figure \ref{fig:router_cat_distribution} shows category distribution for the final training data. 

For solving the category-to-expert mapping a desired completion for each prompt is required. We choose completion as the response from the \emph{GPT-4o-2024-02-15-preview} with goal of constructing a CoE to match the quality of large state-of-art model. 

\begin{figure}
     \centering
    \includegraphics[scale=0.425]{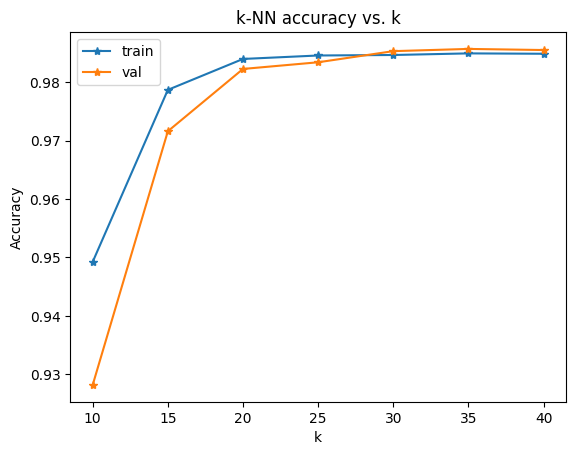}
    \caption{Training validation accuracy vs. $k$ for $k$-NN based category router $\textrm{CR}$ using \textit{intfloat/e5-mistral-7b-instruct} as the text embedding model.}
    \label{fig:train_val_error_plot} 
\end{figure}
\begin{figure}
     \centering
    \includegraphics[scale=0.45]{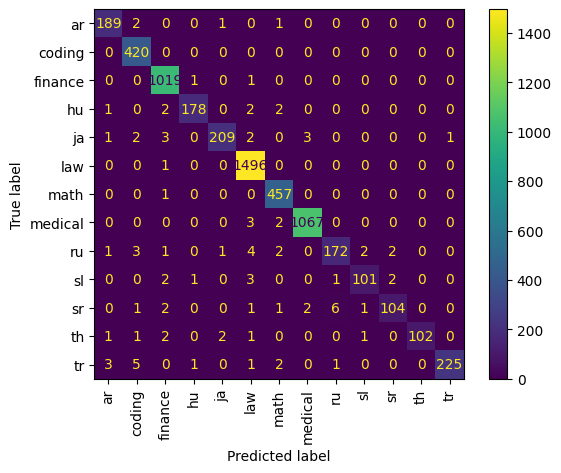}
    \caption{Confusion matrix on the test split for $k$-NN ($k=30$) based the category router $\textrm{CR}$ using \textit{intfloat/e5-mistral-7b-instruct} as the text embedding model.}
    \label{fig:confusion_matrix} 
\end{figure}
\begin{figure*}
     \centering
    \includegraphics[scale=0.45]{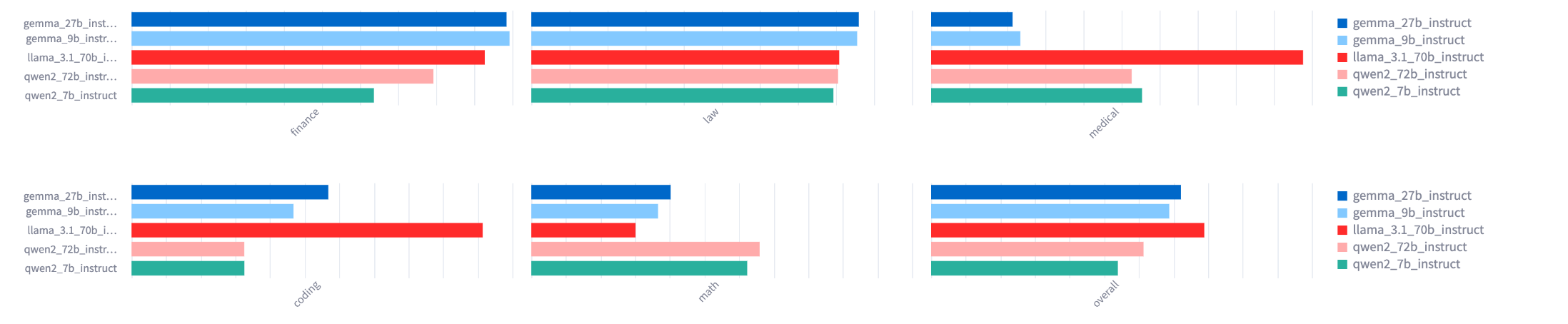}
    \caption{Winrates of expert LLMs \textit{Qwen/Qwen2-7B-Instruct, google/gemma-2-9b-it, google/gemma-2-27b-it, meta-llama/Llama-3.1-70B-Instruct and Qwen/Qwen2-72B-Instruct} on different domains. The win-rates are computed using LLM-as-a-judge approach as described in Section \ref{sec:cat_to_map_training}. We can see that even popular general LLMs rank differently depending on the category.}
    \label{fig:expert_win_rates_domain} 
\end{figure*}
\subsection{Category Router Training}
Figure \ref{fig:2d_tsne_plot_category} shows the 2D t-SNE plot of prompt embeddings obtained from text-embedding model \textit{intfloat/e5-mistral-7b-instruct} and with colors indicating the category label of each prompt. The clustering of different categories into distinct cluster shows that these embeddings from  \textit{intfloat/e5-mistral-7b-instruct} are informative enough for building a highly accurate category router. 
Encouraged by this, we choose category router $\textrm{CR}(\cdot)$  architecture as the $k$-NN classifier built on top of a text-to-embedding backbone. We train the category router on top of \textit{intfloat/e5-mistral-7b-instruct} embeddings and use a simple $k$-NN multi-class classifier.
We split the given training data into train, test and validation splits. We train $k$-NN for various of $k=10, 15, 20, 25, 30, 35, 40$. The router accuracy on training and validation split for various values of $k$ is shown in Figure \ref{fig:train_val_error_plot}. Based on accuracy in validation split we choose $k=30$ for $k$-NN. The router achieves $98\%$ accuracy on the test and the confusion matrix is shown in the Figure \ref{fig:confusion_matrix}. It is quite evident from the confusion matrix that very high accuracy for category router can be achieved.
For the Robust-CoE, we choose entropy obtained from the probabilities that the $k$-NN classifier assigns to various categories. If the measured entropy for a given prompt is larger than the entropy threshold $\epsilon_{t}$ we assign it to 'general' category. We consider various values of $\epsilon_{t} \in \{0.05, 0.10, \cdots, 1.5 \}$ we choose the entropy threshold that leads minimal reduction in the recall of the trained category-router. Based on this, we choose $\epsilon_{t} = 0.1$ for the Robust-CoE.  

\subsection{Category-to-expert mapping training}\label{sec:cat_to_map_training}

\begin{figure}
    \centering
    \includegraphics[scale=0.52]{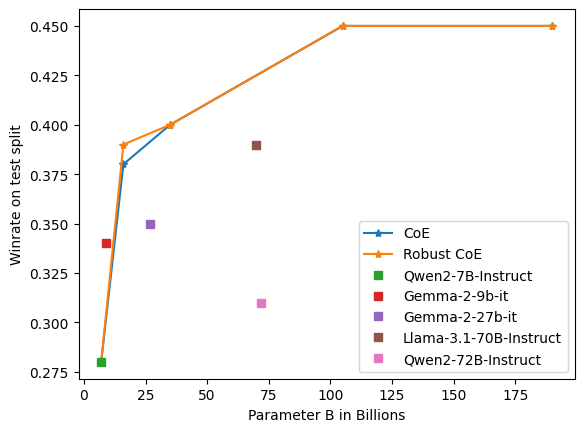}
    \caption{Win-rates against large monolithic model GPT-4o-2024-02-15-preview on on the test split vs. $B$ for experts LLMs, CoE and Robust-CoE. Various CoEs use $k$-NN ($k=30$) based category router using \textit{intfloat/e5-mistral-7b-instruct} as the text embedding model and category-to-expert mapping with parameter budget $B$.}
    \label{fig:CoE_Test_Set_Performance} 
\end{figure}

\begin{figure*}
     \centering
    \includegraphics[scale=0.45]{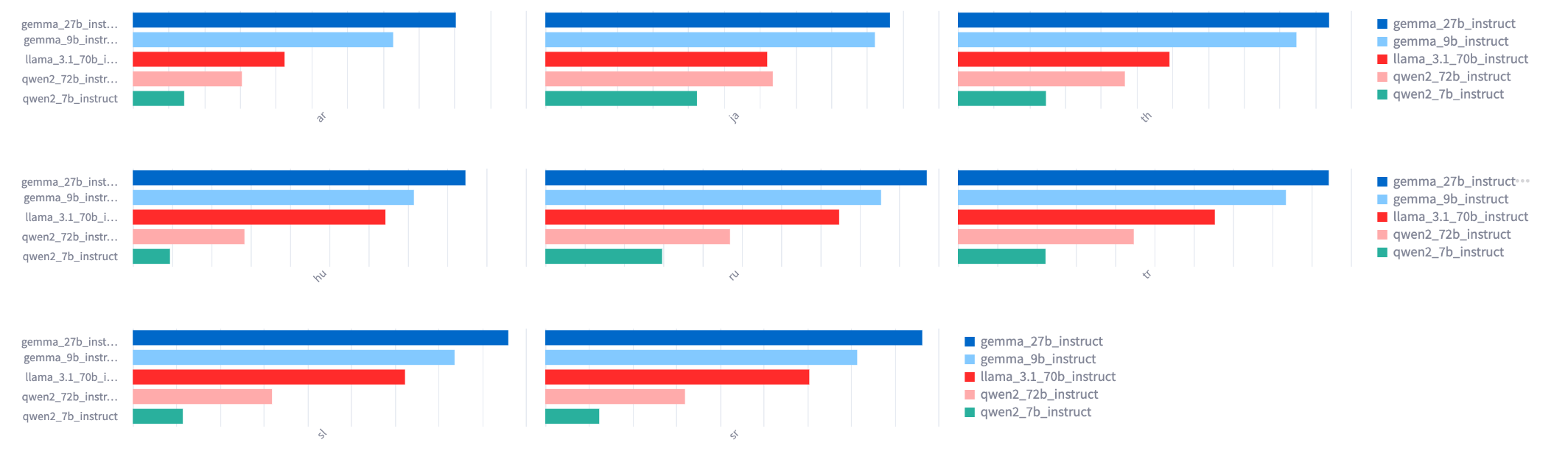}
    \caption{Winrates of expert LLMs \textit{Qwen/Qwen2-7B-Instruct, google/gemma-2-9b-it, google/gemma-2-27b-it, meta-llama/Llama-3.1-70B-Instruct and Qwen/Qwen2-72B-Instruct} on different languages. The win-rates are computed using LLM-as-a-judge approach as described in Section \ref{sec:cat_to_map_training}.}
    \label{fig:expert_win_rates_multilingual} 
\end{figure*}

We obtain, $l_{i,j}$ in category-to-expert mapping optimization problem \eqref{eqn:MILP_Complicated} by computing the win-rate of $j^{th}$ expert on set of prompts routed to category $i^{th}$ using LLM-as-a-judge and equating $l_{i,j}$ as the negative of the win-rate. We use the judge prompt template in \cite{raju2024constructing} given its emphasis on effectively addresses the nuances associated with using LLM as a judge for variety of domains. We use \emph{GPT-4o-mini} as the judge model and generate judgment for expert completion in terms of whether it is similar or better and worse than the desired completion. Equipped with these judgments we can calculate win-rate expert $j$ on category $i$. Figure \ref{fig:expert_win_rates_domain} and \ref{fig:expert_win_rates_multilingual} show that expert win-rates on various languages and domains. We observe that even general mainstream LLMs in our expert set have different capabilities in various categories. This can be attributed to the our training data curation methodology. 

Equipped with $l_{i,j}$, we solve \eqref{eq: MILP} to obtain category-to-expert mapping for various values of parameter budget $B
\in \{7, 16, 35, 105, 190\}$ in billions of parameters. Figure \ref{fig:CoE_Test_Set_Performance} shows performance of CoEs and Robust CoEs for various values of $B$. We observe that the winrates for CoEs w.r.t \emph{GPT-4o-2024-02-15-preview} improves as we increase $B$ and they perform better individual experts. This shows performance w.r.t. large monolithic models can be improved via using CoE constructed out of readily available open source/weight models. We also observe that the performance of both CoE and Robust CoE are similar. This similarity in performance can be attributed to the fact that category-to-expert mapping chooses the best expert for the prompts routed to the `general' category and presence of strong general models in the expert pool. 


\section{Experimental Results}

We evaluate the CoEs for different values of $B \in \{7, 16, 35, 105, 190\}$  on popular LLM Benchmarks such as Arena-Hard \cite{Arena_Hard}, MT-Bench \cite{MT_Bench} and various domain knowledge specific benchmark in MMLU-Pro \cite{MMLU_Pro}. We choose Arena-hard for measuring single-turn evaluations and MT-Bench to measure multi-turn evaluations for CoEs. These benchmarks contain challenging prompts collected in a crowd-sourced manner via Chatbot-Arena and represent a real-world usage of LLMs \cite{chiang2024chatbotarenaopenplatform}. For all the benchmark we measure performance against average number of active parameters for that benchmark. This can be obtained by taking average of the size of the expert chosen by the router for prompts in the benchmark.  We emphasize on measuring average number of parameters because it translates to better tokens per second and time-to-first-token. These some of the key metrics in real-world LLM applications.  
\subsection{Arena-Hard}
\begin{figure}
    \centering
        \includegraphics[scale=0.5]{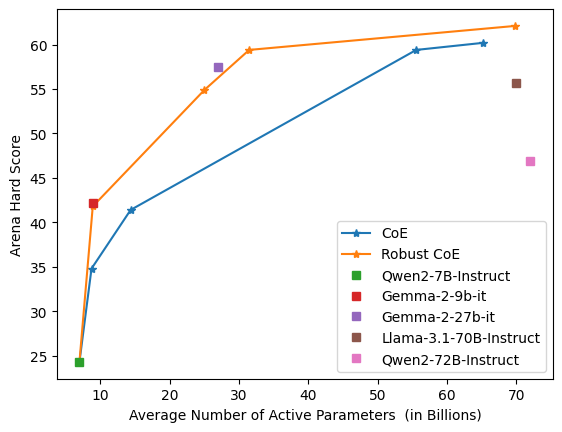}
        \caption{Area-Hard scores vs. average number of active parameters in Billions for expert LLMs, CoEs and Robust-CoEs.}
        \label{fig:arena-hard}
\end{figure}

Figure \ref{fig:arena-hard} shows performance on the Arena-Hard benchmark for CoE for various values of $B$. On the y-axis is the the Arena-Hard score and on the $x$-axis is the average number of active parameters for the benchmark. As total size $B$ of CoE increases, CoE achieves better performance with fewer active performance and is better than individual experts. This shows that the performance of CoE scales as more experts added. We also observe that Robust-CoE significantly outperforms CoE, demonstrating the value of uncertainty quantified routing. It is key to the stable performance of CoE beyond the training dataset. The CoE and Robust-CoE with total parameters $B=190$ Billion achieves the score $60.2$ and $62.10$ with merely $65.23$ and $69.81$ Billion active parameters. Some of popular closed sourced model \emph{claude-3-opus-20240229}, \emph{gemini-1.5-flash-api-preview} and \emph{gpt-4-0613} achieve scores $60.4, 49.6$ and $37.9$ respectively. Due to closed sourced nature of these models, the number of parameters are not known. While the CoE is not state-of-art models on the benchmark, the performance of CoEs can be improved by easily by including new models that being actively developed by community leveraging the modular architecture of CoE.

\subsection{MT-Bench}
\begin{figure}
    \centering
        \includegraphics[scale=0.5]{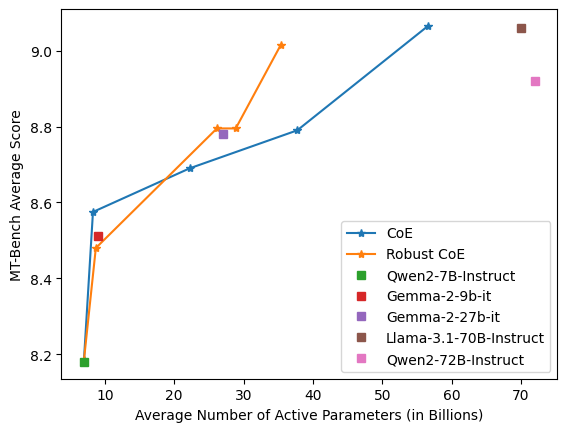}
        \caption{Average MT-Bench scores vs. average number of active parameters in Billions for expert LLMs, CoEs and Robust-CoEs.} \label{fig:MT_Bench_Average}
        \label{fig:MT_Bench_Results}
\end{figure}

The performance on various metrics on MT-Bench and the average number of active parameters in the CoE are shown in Figure \ref{fig:MT_Bench_Results}. This benchmark measures performance of LLMs on multi-turn conversation. The CoE routes based on current input and then passes the entire conversation history. This means different experts might be chosen for different conversation turns.
Figure \ref{fig:MT_Bench_Average} shows the average score over multi-turn conversations in the benchmark and average number of active parameters. They achieve similar performance to large expert models \emph{meta-llama/Llama-3.1-70B-Instruct} and \emph{Qwen/Qwen2-72B-Instruct}  at a much reduced average active parameters. 

\subsection{Knowledge Intensive Benchmarks}
\begin{figure}
    \centering
    \includegraphics[scale=0.5]{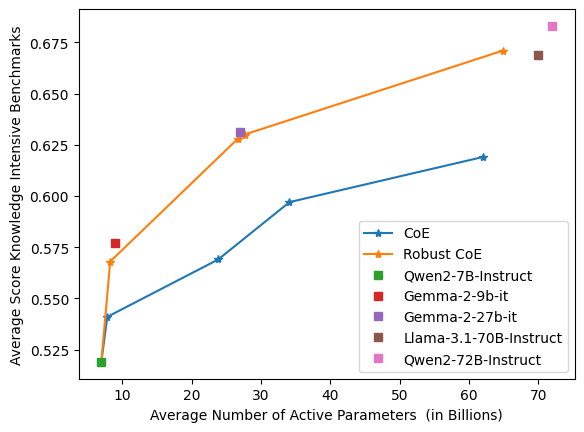}
    \caption{Average scores on knowledge intensive benchmarks vs. average number of active parameters in Billions for expert LLMs, CoEs and Robust-CoEs.}
    \label{fig:Knowledge Intensive}
\end{figure}

We measure the performance of CoE on knowledge intensive benchmarks such GSM8k CoT and MMLU-pro benchmark on the following subjects:  Mathematics, Business, Computer Science, Economics, Health and Law. The average score on these benchmarks vs. average number of active parameters are shown in Figure \ref{fig:Knowledge Intensive}.  We observe that while CoE does suffer from reduced performance as compared to individual experts, the Robust-CoE recovers most of this performance.  The fine-grained results are shown in Table \ref{tab:kt_benchmark_results}.  This can be attributed to the fact that the distribution mismatch between knowledge intensive tasks and the router training data. The robust-CoE via uncertainty quantification detects this mismatch and appropriately routes these prompts to the best generalist expert.

%% file: CoE_Appendix.tex
\appendix


\section{Methodology to create multilingual prompts} \label{sec:multilingual_category}
Multilingual prompts can be categorized into three overarching formats: native, translation, and cross-lingual. In native formats, prompts are written in the target language, and the model is expected to respond in the same language. Translation formats involve queries requesting translation either into or from a specific language. Cross-lingual formats contain prompts written in one language while explicitly requesting responses in another language, such as "Please say hello in Arabic." 
\begin{table*}[htb!]
    \centering
    \small
    \begin{tabular}{|c|c|}
    \hline
        \textbf{Domain} & \textbf{HuggingFace Dataset Name} \\
    \hline
      Medical   &  ruslanmv/ai-medical-chatbot  \\
      & qiaojin/PubMedQA (pqd\_labeled subset) \\
      \hline
      Finance   & gbharti/finance-alpaca \\
      & BeIR/fiqa queries \\
      & nihiluis/financial-advisor-100 \\
      \hline
      Coding & jondurbin/airoboros-2.1  \\
      \hline
      Mathematics   &    The following sources from allenai/lila: \\
      & MATH\_algebra\_crowdsourced  \\
    &  MATH\_counting\_and\_probability\_crowdsourced \\ 
    & MATH\_intermediate\_algebra\_crowdsourced \\
    &  mathqa\_gain, mathqa\_general, mathqa\_geometry \\
    &  mathqa\_other, mathqa\_physics, mathqa\_probability \\
      \hline
      Law & jonathanli/legal-advice-reddit \\
      & lighteval/legal\_summarization (BillSum subset)   \\
      \hline
    \end{tabular}
    \caption{Source dataset for each domain.}
    \label{tab:domain_dataset_sources}
\end{table*}
For native prompts, we sampled data from several datasets: CohereForAI/aya\_dataset \cite{singh2024ayadatasetopenaccesscollection}, allenai/WildChat-1M \cite{zhao2024wildchat1mchatgptinteraction}, OpenAssistant/oasst1 \cite{köpf2023openassistant}, and sambanovasystems/x-self-instruct-seed-32 \cite{xinstruct}. We filtered and sorted the data by language using the language labels provided in these datasets.

Upon analysis, we found that the native prompts were primarily categorized by Language Identification systems (LIDs), which excluded translation prompts and English prompts requesting responses in other languages. To address this limitation, we generated synthetic prompts for both translation and cross-lingual scenarios. For translation prompts, we began by manually creating five diverse template examples requesting translation from the target language to English. These templates were then used as few-shot examples for an LLM to generate 100 additional diverse translation request templates. We applied these templates to the FLORES-200 dataset \citep{flores} to create our synthetic translation dataset. The process was then repeated for generating prompts requesting translation from English to the target language. An example template is ``Could you convert \{query\} from \{in-lang\} to \{out-lang\}?". For cross-lingual prompts, we followed a similar approach, creating five unique templates and using an LLM to generate 100 more based on these examples. These templates were ten applied to the UltraChat dataset queries \cite{cui2023ultrafeedback}. An example template is ``Please write your response in \{lang\} to: \{query\}."

\begin{table*}[htb!]
    \centering
    \tiny
    \begin{tabular}{|c|c|c|c|c|c|c|c|c|c|} 
    \hline
       Model & GSM8k CoT & MMLU Pro   &  MMLU Pro   & MMLU Pro  & MMLU Pro  &  MMLU Pro  & MMLU Pro &  MMLU Pro & Average   \\
         & (Flexible Match) & Biology  &  Business  & Computer Science   &  Economics  &   Health & Law &  Math &  Score \\
    \hline
       Qwen2-7B-Instruct  &   0.780 & 	0.648 &	0.512 &	0.448 &	0.581 &	0.465 &	0.299	& 0.419	& 0.519\\
    \hline
        gemma-2-9b-it  &	0.816	& 0.768	& 0.523	& 0.487& 	0.643	& 0.592 &	0.346 &	0.441	& 0.577 \\
    \hline
        gemma-2-27b-it   &	0.844 &	0.785&	0.556	&0.597&	0.699	& 0.647&	0.415&	0.507&	0.631 \\
    \hline
       Llama-3.1-70B-Instruct 	& 0.924 &	0.779 &	0.618 &	0.629 &	0.738&	0.657 &	0.458 &	0.547 &	0.669\\
    \hline
        Qwen2-72B-Instruct  & 0.901 &	0.813 &	0.637	 & 0.639 & 0.763	& 0.667	 & 0.423	& 0.620 &	0.683\\
    \hline
       CoE 7B & 0.780 & 0.648 & 0.512 & 0.448 & 0.581 & 0.465 & 0.299 & 0.419 & 0.519 \\
       \hline
       CoE 16B & 0.780	 & 0.695 &	0.522 &	0.463	&0.643&	0.459&	0.344&	0.419&	0.541 \\
       \hline
       CoE 35B &  0.844 & 0.739 &	0.448 &	0.568&	0.522&	0.505	&0.415&	0.507	&0.569 \\
       \hline
       CoE 105B &  0.780 &	0.726 & 0.556&	0.514&	0.700&	0.664	&0.415&	0.419	&0.597 \\
       \hline
       CoE 190B & 0.901 &	0.781 & 0.447&	0.604	&0.523	&0.664	&0.415	&0.620	&0.619 \\
       \hline
      Robust CoE 7B & 0.780 & 0.648 & 0.512 & 0.448 & 0.581 & 0.465 & 0.299 & 0.419 & 0.519  \\
       \hline
      Robust CoE 16B & 0.780 &	0.767 &	0.523 &	0.487 &	0.643 &	0.577 &	0.344 &	0.423 &	0.568 \\
       \hline
      Robust CoE 35B & 0.844 &	0.785 &	0.556 &	0.597 &	0.699 &	0.624 &	0.415 &	0.507 &	0.628\\
       \hline
      Robust CoE 105B & 0.844 & 0.785	 & 0.556 &	0.597 &	0.699 &	0.637 &	0.415 &	0.507 &	0.630 \\
       \hline
      Robust CoE 190B & 0.901	& 0.781	& 0.618	& 0.626 &	0.738 &	0.664 &	0.422 &	0.620 &	0.671 \\
       \hline    
    \end{tabular}
    \newline
    \caption{\small Performance on knowledge intensive domain specific benchmarks.}
    \label{tab:kt_benchmark_results}
\end{table*}

%% file: Main.bbl
\begin{thebibliography}{45}
\providecommand{\natexlab}[1]{#1}
\providecommand{\url}[1]{\texttt{#1}}
\expandafter\ifx\csname urlstyle\endcsname\relax
  \providecommand{\doi}[1]{doi: #1}\else
  \providecommand{\doi}{doi: \begingroup \urlstyle{rm}\Url}\fi

\bibitem[aa()]{aa}
Artificial analysis: Independent analysis of ai models and api providers.
\newblock \url{https://artificialanalysis.ai/}.

\bibitem[dgx()]{dgx-h100}
Nvidia dgx h100 datasheet.
\newblock \url{https://resources.nvidia.com/en-us-dgx-systems/ai-enterprise-dgx}.

\bibitem[gh2()]{gh200}
Nvidia dgx gh200 datasheet.
\newblock \url{https://resources.nvidia.com/en-us-dgx-gh200/nvidia-dgx-gh200-datasheet-web-us}.

\bibitem[sam()]{samba1}
Benchmarking samba-1.
\newblock \url{https://sambanova.ai/blog/benchmarking-samba-1}.

\bibitem[Chiang et~al.(2024)Chiang, Zheng, Sheng, Angelopoulos, Li, Li, Zhang, Zhu, Jordan, Gonzalez, and Stoica]{chiang2024chatbotarenaopenplatform}
Chiang, W.-L., Zheng, L., Sheng, Y., Angelopoulos, A.~N., Li, T., Li, D., Zhang, H., Zhu, B., Jordan, M., Gonzalez, J.~E., and Stoica, I.
\newblock Chatbot arena: An open platform for evaluating llms by human preference, 2024.
\newblock URL \url{https://arxiv.org/abs/2403.04132}.

\bibitem[Cui et~al.(2023)Cui, Yuan, Ding, Yao, Zhu, Ni, Xie, Liu, and Sun]{cui2023ultrafeedback}
Cui, G., Yuan, L., Ding, N., Yao, G., Zhu, W., Ni, Y., Xie, G., Liu, Z., and Sun, M.
\newblock Ultrafeedback: Boosting language models with high-quality feedback, 2023.

\bibitem[Goodfellow et~al.(2018)Goodfellow, McDaniel, and Papernot]{goodfellow2018making}
Goodfellow, I., McDaniel, P., and Papernot, N.
\newblock Making machine learning robust against adversarial inputs.
\newblock \emph{Communications of the ACM}, 61\penalty0 (7):\penalty0 56--66, 2018.

\bibitem[Goyal et~al.(2021)Goyal, Gao, Chaudhary, Chen, Wenzek, Ju, Krishnan, Ranzato, Guzm\'{a}n, and Fan]{flores}
Goyal, N., Gao, C., Chaudhary, V., Chen, P.-J., Wenzek, G., Ju, D., Krishnan, S., Ranzato, M., Guzm\'{a}n, F., and Fan, A.
\newblock The flores-101 evaluation benchmark for low-resource and multilingual machine translation.
\newblock 2021.

\bibitem[Gururangan et~al.(2023)Gururangan, Li, Lewis, Shi, Althoff, Smith, and Zettlemoyer]{gururangan2023scaling}
Gururangan, S., Li, M., Lewis, M., Shi, W., Althoff, T., Smith, N.~A., and Zettlemoyer, L.
\newblock Scaling expert language models with unsupervised domain discovery, 2023.

\bibitem[He et~al.(2016)He, Zhang, Ren, and Sun]{he2016deep}
He, K., Zhang, X., Ren, S., and Sun, J.
\newblock Deep residual learning for image recognition.
\newblock In \emph{Proceedings of the IEEE conference on computer vision and pattern recognition}, pp.\  770--778, 2016.

\bibitem[Jang et~al.(2023)Jang, Kim, Ye, Kim, Logeswaran, Lee, Lee, and Seo]{jang2023exploring}
Jang, J., Kim, S., Ye, S., Kim, D., Logeswaran, L., Lee, M., Lee, K., and Seo, M.
\newblock Exploring the benefits of training expert language models over instruction tuning, 2023.

\bibitem[Jiang et~al.(2023{\natexlab{a}})Jiang, Sablayrolles, Mensch, Bamford, Chaplot, Casas, Bressand, Lengyel, Lample, Saulnier, et~al.]{jiang2023mistral}
Jiang, A.~Q., Sablayrolles, A., Mensch, A., Bamford, C., Chaplot, D.~S., Casas, D. d.~l., Bressand, F., Lengyel, G., Lample, G., Saulnier, L., et~al.
\newblock Mistral 7b.
\newblock \emph{arXiv preprint arXiv:2310.06825}, 2023{\natexlab{a}}.

\bibitem[Jiang et~al.(2024)Jiang, Sablayrolles, Roux, Mensch, Savary, Bamford, Chaplot, Casas, Hanna, Bressand, et~al.]{jiang2024mixtral}
Jiang, A.~Q., Sablayrolles, A., Roux, A., Mensch, A., Savary, B., Bamford, C., Chaplot, D.~S., Casas, D. d.~l., Hanna, E.~B., Bressand, F., et~al.
\newblock Mixtral of experts.
\newblock \emph{arXiv preprint arXiv:2401.04088}, 2024.

\bibitem[Jiang et~al.(2023{\natexlab{b}})Jiang, Ren, and Lin]{jiang2023llm}
Jiang, D., Ren, X., and Lin, B.~Y.
\newblock Llm-blender: Ensembling large language models with pairwise ranking and generative fusion.
\newblock \emph{arXiv preprint arXiv:2306.02561}, 2023{\natexlab{b}}.

\bibitem[Ju et~al.(2018)Ju, Bibaut, and van~der Laan]{ju2018relative}
Ju, C., Bibaut, A., and van~der Laan, M.
\newblock The relative performance of ensemble methods with deep convolutional neural networks for image classification.
\newblock \emph{Journal of Applied Statistics}, 45\penalty0 (15):\penalty0 2800--2818, 2018.

\bibitem[Kaplan et~al.(2020)Kaplan, McCandlish, Henighan, Brown, Chess, Child, Gray, Radford, Wu, and Amodei]{kaplan2020scaling}
Kaplan, J., McCandlish, S., Henighan, T., Brown, T.~B., Chess, B., Child, R., Gray, S., Radford, A., Wu, J., and Amodei, D.
\newblock Scaling laws for neural language models.
\newblock \emph{arXiv preprint arXiv:2001.08361}, 2020.

\bibitem[Kumar \& Byrne(2004)Kumar and Byrne]{kumar2004minimum}
Kumar, S. and Byrne, B.
\newblock Minimum bayes-risk decoding for statistical machine translation.
\newblock In \emph{Proceedings of the Human Language Technology Conference of the North American Chapter of the Association for Computational Linguistics: HLT-NAACL 2004}, pp.\  169--176, 2004.

\bibitem[Köpf et~al.(2023)Köpf, Kilcher, von Rütte, Anagnostidis, Tam, Stevens, Barhoum, Duc, Stanley, Nagyfi, ES, Suri, Glushkov, Dantuluri, Maguire, Schuhmann, Nguyen, and Mattick]{köpf2023openassistant}
Köpf, A., Kilcher, Y., von Rütte, D., Anagnostidis, S., Tam, Z.-R., Stevens, K., Barhoum, A., Duc, N.~M., Stanley, O., Nagyfi, R., ES, S., Suri, S., Glushkov, D., Dantuluri, A., Maguire, A., Schuhmann, C., Nguyen, H., and Mattick, A.
\newblock Openassistant conversations -- democratizing large language model alignment, 2023.

\bibitem[Lepikhin et~al.(2020)Lepikhin, Lee, Xu, Chen, Firat, Huang, Krikun, Shazeer, and Chen]{lepikhin2020gshard}
Lepikhin, D., Lee, H., Xu, Y., Chen, D., Firat, O., Huang, Y., Krikun, M., Shazeer, N., and Chen, Z.
\newblock Gshard: Scaling giant models with conditional computation and automatic sharding.
\newblock \emph{arXiv preprint arXiv:2006.16668}, 2020.

\bibitem[Li et~al.(2022)Li, Gururangan, Dettmers, Lewis, Althoff, Smith, and Zettlemoyer]{li2022branchtrainmerge}
Li, M., Gururangan, S., Dettmers, T., Lewis, M., Althoff, T., Smith, N.~A., and Zettlemoyer, L.
\newblock Branch-train-merge: Embarrassingly parallel training of expert language models, 2022.

\bibitem[Li et~al.(2024)Li, Chiang, Frick, Dunlap, Wu, Zhu, Gonzalez, and Stoica]{Arena_Hard}
Li, T., Chiang, W.-L., Frick, E., Dunlap, L., Wu, T., Zhu, B., Gonzalez, J.~E., and Stoica, I.
\newblock From crowdsourced data to high-quality benchmarks: Arena-hard and benchbuilder pipeline, 2024.
\newblock URL \url{https://arxiv.org/abs/2406.11939}.

\bibitem[lin Chiang et~al.(2024)lin Chiang, Zheng, Dunlap, Gonzalez, Stoica, Mooney, Dane, Howard, and Keating]{lmsys-chatbot-arena}
lin Chiang, W., Zheng, L., Dunlap, L., Gonzalez, J.~E., Stoica, I., Mooney, P., Dane, S., Howard, A., and Keating, N.
\newblock Lmsys - chatbot arena human preference predictions.
\newblock \url{https://kaggle.com/competitions/lmsys-chatbot-arena}, 2024.
\newblock Kaggle.

\bibitem[Lu et~al.(2023)Lu, Yuan, Lin, Lin, Yuan, Zhou, and Zhou]{lu2023routing}
Lu, K., Yuan, H., Lin, R., Lin, J., Yuan, Z., Zhou, C., and Zhou, J.
\newblock Routing to the expert: Efficient reward-guided ensemble of large language models.
\newblock \emph{arXiv preprint arXiv:2311.08692}, 2023.

\bibitem[Luo et~al.(2024)Luo, Yang, Meng, Li, Zhou, and Zhang]{luo2024empiricalstudycatastrophicforgetting}
Luo, Y., Yang, Z., Meng, F., Li, Y., Zhou, J., and Zhang, Y.
\newblock An empirical study of catastrophic forgetting in large language models during continual fine-tuning, 2024.
\newblock URL \url{https://arxiv.org/abs/2308.08747}.

\bibitem[Muennighoff et~al.(2023)Muennighoff, Tazi, Magne, and Reimers]{muennighoff2023mtebmassivetextembedding}
Muennighoff, N., Tazi, N., Magne, L., and Reimers, N.
\newblock Mteb: Massive text embedding benchmark, 2023.
\newblock URL \url{https://arxiv.org/abs/2210.07316}.

\bibitem[Ouyang et~al.(2022)Ouyang, Wu, Jiang, Almeida, Wainwright, Mishkin, Zhang, Agarwal, Slama, Ray, et~al.]{ouyang2022training}
Ouyang, L., Wu, J., Jiang, X., Almeida, D., Wainwright, C., Mishkin, P., Zhang, C., Agarwal, S., Slama, K., Ray, A., et~al.
\newblock Training language models to follow instructions with human feedback.
\newblock \emph{Advances in Neural Information Processing Systems}, 35:\penalty0 27730--27744, 2022.

\bibitem[Prabhakar(2024)]{sn40l-hotchips}
Prabhakar, R.
\newblock Sambanova sn40l rdu: Breaking the barrier of trillion+ parameter scale gen ai computing.
\newblock In \emph{2024 IEEE Hot Chips 36 Symposium (HCS)}, pp.\  1--24, Los Alamitos, CA, USA, aug 2024. IEEE Computer Society.
\newblock \doi{10.1109/HCS61935.2024.10664717}.
\newblock URL \url{https://doi.ieeecomputersociety.org/10.1109/HCS61935.2024.10664717}.

\bibitem[Prabhakar et~al.(2024)Prabhakar, Sivaramakrishnan, Gandhi, Du, Wang, Song, Zhang, Gao, Wang, Li, Sheng, Brot, Sokolov, Vivek, Leung, Sabnis, Bai, Zhao, Gottscho, Jackson, Luttrell, Shah, Chen, Liang, Jain, Thakker, Huang, Jairath, Brown, and Olukotun]{prabhakar2024sambanova}
Prabhakar, R., Sivaramakrishnan, R., Gandhi, D., Du, Y., Wang, M., Song, X., Zhang, K., Gao, T., Wang, A., Li, K., Sheng, Y., Brot, J., Sokolov, D., Vivek, A., Leung, C., Sabnis, A., Bai, J., Zhao, T., Gottscho, M., Jackson, D., Luttrell, M., Shah, M.~K., Chen, E., Liang, K., Jain, S., Thakker, U., Huang, D., Jairath, S., Brown, K.~J., and Olukotun, K.
\newblock Sambanova sn40l: Scaling the ai memory wall with dataflow and composition of experts, 2024.
\newblock URL \url{https://arxiv.org/abs/2405.07518}.

\bibitem[Raju et~al.(2024)Raju, Jain, Li, Li, and Thakkar]{raju2024constructing}
Raju, R., Jain, S., Li, B., Li, J., and Thakkar, U.
\newblock Constructing domain-specific evaluation sets for llm-as-a-judge.
\newblock \emph{arXiv preprint arXiv:2408.08808}, 2024.

\bibitem[Rakesh \& Jain(2021)Rakesh and Jain]{rakesh2021efficacy}
Rakesh, V. and Jain, S.
\newblock Efficacy of bayesian neural networks in active learning.
\newblock In \emph{Proceedings of the IEEE/CVF Conference on Computer Vision and Pattern Recognition}, pp.\  2601--2609, 2021.

\bibitem[Settles(2009)]{settles2009active}
Settles, B.
\newblock Active learning literature survey.
\newblock Computer Sciences Technical Report 1648, University of Wisconsin--Madison, 2009.
\newblock URL \url{http://axon.cs.byu.edu/~martinez/classes/778/Papers/settles.activelearning.pdf}.

\bibitem[Sheikholeslami et~al.(2019)Sheikholeslami, Jain, and Giannakis]{sheikholeslami2019efficient}
Sheikholeslami, F., Jain, S., and Giannakis, G.~B.
\newblock Efficient randomized defense against adversarial attacks in deep convolutional neural networks.
\newblock In \emph{ICASSP 2019-2019 IEEE International Conference on Acoustics, Speech and Signal Processing (ICASSP)}, pp.\  3277--3281. IEEE, 2019.

\bibitem[Sheikholeslami et~al.(2020)Sheikholeslami, Jain, and Giannakis]{sheikholeslami2020minimum}
Sheikholeslami, F., Jain, S., and Giannakis, G.~B.
\newblock Minimum uncertainty based detection of adversaries in deep neural networks.
\newblock In \emph{2020 Information Theory and Applications Workshop (ITA)}, pp.\  1--16. IEEE, 2020.

\bibitem[Shnitzer et~al.(2023)Shnitzer, Ou, Silva, Soule, Sun, Solomon, Thompson, and Yurochkin]{shnitzer2023llm}
Shnitzer, T., Ou, A., Silva, M., Soule, K., Sun, Y., Solomon, J., Thompson, N., and Yurochkin, M.
\newblock Llm routing with benchmark datasets.
\newblock In \emph{NeurIPS 2023 Workshop on Distribution Shifts: New Frontiers with Foundation Models}, 2023.

\bibitem[Singh et~al.(2024)Singh, Vargus, Dsouza, Karlsson, Mahendiran, Ko, Shandilya, Patel, Mataciunas, OMahony, Zhang, Hettiarachchi, Wilson, Machado, Moura, Krzemiński, Fadaei, Ergün, Okoh, Alaagib, Mudannayake, Alyafeai, Chien, Ruder, Guthikonda, Alghamdi, Gehrmann, Muennighoff, Bartolo, Kreutzer, Üstün, Fadaee, and Hooker]{singh2024ayadatasetopenaccesscollection}
Singh, S., Vargus, F., Dsouza, D., Karlsson, B.~F., Mahendiran, A., Ko, W.-Y., Shandilya, H., Patel, J., Mataciunas, D., OMahony, L., Zhang, M., Hettiarachchi, R., Wilson, J., Machado, M., Moura, L.~S., Krzemiński, D., Fadaei, H., Ergün, I., Okoh, I., Alaagib, A., Mudannayake, O., Alyafeai, Z., Chien, V.~M., Ruder, S., Guthikonda, S., Alghamdi, E.~A., Gehrmann, S., Muennighoff, N., Bartolo, M., Kreutzer, J., Üstün, A., Fadaee, M., and Hooker, S.
\newblock Aya dataset: An open-access collection for multilingual instruction tuning, 2024.
\newblock URL \url{https://arxiv.org/abs/2402.06619}.

\bibitem[Sukhbaatar et~al.(2024)Sukhbaatar, Golovneva, Sharma, Xu, Lin, Rozière, Kahn, Li, tau Yih, Weston, and Li]{sukhbaatar2024branchtrainmix}
Sukhbaatar, S., Golovneva, O., Sharma, V., Xu, H., Lin, X.~V., Rozière, B., Kahn, J., Li, D., tau Yih, W., Weston, J., and Li, X.
\newblock Branch-train-mix: Mixing expert llms into a mixture-of-experts llm, 2024.

\bibitem[Systems(2023)]{xinstruct}
Systems, S.
\newblock x-self-instruct-seed-32, 5 2023.
\newblock URL \url{https://huggingface.co/datasets/sambanovasystems/x-self-instruct-seed-32}.

\bibitem[Team et~al.(2024)Team, Riviere, Pathak, Sessa, Hardin, Bhupatiraju, Hussenot, Mesnard, Shahriari, Ram{\'e}, et~al.]{team2024gemma}
Team, G., Riviere, M., Pathak, S., Sessa, P.~G., Hardin, C., Bhupatiraju, S., Hussenot, L., Mesnard, T., Shahriari, B., Ram{\'e}, A., et~al.
\newblock Gemma 2: Improving open language models at a practical size.
\newblock \emph{arXiv preprint arXiv:2408.00118}, 2024.

\bibitem[Touvron et~al.(2023)Touvron, Martin, Stone, Albert, Almahairi, Babaei, Bashlykov, Batra, Bhargava, Bhosale, et~al.]{touvron2023llama2}
Touvron, H., Martin, L., Stone, K., Albert, P., Almahairi, A., Babaei, Y., Bashlykov, N., Batra, S., Bhargava, P., Bhosale, S., et~al.
\newblock Llama 2: Open foundation and fine-tuned chat models.
\newblock \emph{arXiv preprint arXiv:2307.09288}, 2023.

\bibitem[Van~der Maaten \& Hinton(2008)Van~der Maaten and Hinton]{van2008visualizing}
Van~der Maaten, L. and Hinton, G.
\newblock Visualizing data using t-sne.
\newblock \emph{Journal of machine learning research}, 9\penalty0 (11), 2008.

\bibitem[Virtanen et~al.(2020)Virtanen, Gommers, Oliphant, Haberland, Reddy, Cournapeau, Burovski, Peterson, Weckesser, Bright, {van der Walt}, Brett, Wilson, Millman, Mayorov, Nelson, Jones, Kern, Larson, Carey, Polat, Feng, Moore, {VanderPlas}, Laxalde, Perktold, Cimrman, Henriksen, Quintero, Harris, Archibald, Ribeiro, Pedregosa, {van Mulbregt}, and {SciPy 1.0 Contributors}]{Scipy}
Virtanen, P., Gommers, R., Oliphant, T.~E., Haberland, M., Reddy, T., Cournapeau, D., Burovski, E., Peterson, P., Weckesser, W., Bright, J., {van der Walt}, S.~J., Brett, M., Wilson, J., Millman, K.~J., Mayorov, N., Nelson, A. R.~J., Jones, E., Kern, R., Larson, E., Carey, C.~J., Polat, {\.I}., Feng, Y., Moore, E.~W., {VanderPlas}, J., Laxalde, D., Perktold, J., Cimrman, R., Henriksen, I., Quintero, E.~A., Harris, C.~R., Archibald, A.~M., Ribeiro, A.~H., Pedregosa, F., {van Mulbregt}, P., and {SciPy 1.0 Contributors}.
\newblock {{SciPy} 1.0: Fundamental Algorithms for Scientific Computing in Python}, 2020.

\bibitem[Wang et~al.(2024)Wang, Ma, Zhang, Ni, Chandra, Guo, Ren, Arulraj, He, Jiang, Li, Ku, Wang, Zhuang, Fan, Yue, and Chen]{MMLU_Pro}
Wang, Y., Ma, X., Zhang, G., Ni, Y., Chandra, A., Guo, S., Ren, W., Arulraj, A., He, X., Jiang, Z., Li, T., Ku, M., Wang, K., Zhuang, A., Fan, R., Yue, X., and Chen, W.
\newblock Mmlu-pro: A more robust and challenging multi-task language understanding benchmark (published at neurips 2024 track datasets and benchmarks), 2024.
\newblock URL \url{https://arxiv.org/abs/2406.01574}.

\bibitem[Zhao et~al.(2024{\natexlab{a}})Zhao, Wang, Abid, Angus, Garg, Kinnison, Sherstinsky, Molino, Addair, and Rishi]{zhao2024lora}
Zhao, J., Wang, T., Abid, W., Angus, G., Garg, A., Kinnison, J., Sherstinsky, A., Molino, P., Addair, T., and Rishi, D.
\newblock Lora land: 310 fine-tuned llms that rival gpt-4, a technical report, 2024{\natexlab{a}}.

\bibitem[Zhao et~al.(2024{\natexlab{b}})Zhao, Ren, Hessel, Cardie, Choi, and Deng]{zhao2024wildchat1mchatgptinteraction}
Zhao, W., Ren, X., Hessel, J., Cardie, C., Choi, Y., and Deng, Y.
\newblock Wildchat: 1m chatgpt interaction logs in the wild, 2024{\natexlab{b}}.
\newblock URL \url{https://arxiv.org/abs/2405.01470}.

\bibitem[Zheng et~al.(2023)Zheng, Chiang, Sheng, Zhuang, Wu, Zhuang, Lin, Li, Li, Xing, Zhang, Gonzalez, and Stoica]{MT_Bench}
Zheng, L., Chiang, W.-L., Sheng, Y., Zhuang, S., Wu, Z., Zhuang, Y., Lin, Z., Li, Z., Li, D., Xing, E.~P., Zhang, H., Gonzalez, J.~E., and Stoica, I.
\newblock Judging llm-as-a-judge with mt-bench and chatbot arena, 2023.
\newblock URL \url{https://arxiv.org/abs/2306.05685}.

\end{thebibliography}
